\pdfoutput=1
\documentclass[11pt]{article}

\usepackage{acl}

\usepackage{times}
\usepackage{latexsym}
\usepackage[T1]{fontenc}
\usepackage{graphicx}

\usepackage[utf8]{inputenc}

\usepackage{microtype}
\usepackage{inconsolata}

\usepackage{tikz}
\usepackage{pgfplots}
\usepackage{algorithm}
\usepackage{algorithmic}
\usepackage{multirow}
\usepackage{tabularx}

\pgfplotsset{compat=1.18}
\usepgfplotslibrary{groupplots}

\usetikzlibrary{external, shapes.geometric, arrows, positioning, matrix, fit}
\tikzstyle{block} = [rectangle, draw, text width=4.9cm, text centered, minimum height=2em]
\tikzstyle{arrow} = [thick, ->, >=stealth]

\title{Detecting Sockpuppetry on Wikipedia Using Meta-Learning}

\author{
    Luc Raszewski \\
    The University of Melbourne \\
    \texttt{lraszewski@student.unimelb.edu.au} \\
    \And
    Christine De Kock \\
    The University of Melbourne \\
    \texttt{christine.dekock@unimelb.edu.au} \\
}

\begin{document}

\maketitle

\begin{abstract}

Malicious sockpuppet detection on Wikipedia is critical to preserving access to reliable information on the internet and preventing the spread of disinformation. Prior machine learning approaches rely on stylistic and meta-data features, but do not prioritise adaptability to author-specific behaviours. As a result, they struggle to effectively model the behaviour of specific sockpuppet-groups, especially when text data is limited. To address this, we propose the application of meta-learning, a machine learning technique designed to improve performance in data-scarce settings by training models across multiple tasks. Meta-learning optimises a model for rapid adaptation to the writing style of a new sockpuppet-group.
Our results show that meta-learning significantly enhances the precision of predictions compared to pre-trained models, marking an advancement in combating sockpuppetry on open editing platforms. We release a new dataset of sockpuppet investigations to foster future research in both sockpuppetry and meta-learning fields.

\end{abstract}
\section{Introduction}

Over recent years, social media sites have seen a steady increase in the presence of fake accounts \citep{khaled2018}. These accounts are often used to spread fake news and seed distrust for political gain \citep{shu2017}. Wikipedia is not immune to such attacks: \citet{saeztrumper2019} investigated political and religious groups imposing their narratives on articles. Attacks on Wikipedia are particularly threatening as articles often serve as the ground-truth for automated fact checking systems; used to combat disinformation on other platforms \citep{thorne2018}.

Changes to articles on Wikipedia are a collaborative process, where decisions are made via the consensus of editors. Malicious users undermine this process through \textit{Sockpuppetry}: the use of multiple accounts to stack votes, fake majority support of a view or make a counter-perspective look absurd \citep{saeztrumper2019}. They have been used to vandalise articles \citep{solorio2013case}, support political views \citep{kumar2017} or improve personal standing \citep{stone2007}.
 
Many machine learning approaches have been proposed, including linking sockpuppet accounts through their writing style \citep{solorio2013case, sakib2022} using a related technique called \textit{authorship-attribution}. However, when the available text is scarce, it is difficult to profile an author accurately \citep{eder2013}. In the case of sockpuppet detection, where available text samples are short \citep{solorio2013sockpuppet}, this makes achieving good performance difficult. Previous approaches \citep{solorio2013case, sakib2022} manage this challenge by merging the corpus of individual sockpuppet investigations into a single dataset. A model trained on this dataset then learns the writing style of sockpuppets as a whole. Whilst the model can be later fine-tuned, it will not be sensitive to author-specific features.

\textit{Meta-learning} instead leverages prior experience to perform well on limited data. Rather than merging the corpus of sockpuppet investigations together and training one large model, it considers each as a separate learning task -- a new model is trained for each individual investigation. These \textit{base-models} begin as a copy of a \textit{meta-model}, a single model that is updated by what is learned by each base-model. The meta-model learns a general understanding of sockpuppets that can quickly adapt to the behaviour of an unseen \textit{puppetmaster}, the user behind a group of sockpuppets. We elaborate on this process in Section \ref{method}.

In this study, we are the first to apply meta-learning to the problem of sockpuppet detection. Our work makes three main contributions:

\paragraph{1. We evaluate the application of meta-learning to the task of sockpuppet detection on Wikipedia.} We find that learning over a distribution of tasks significantly improves prediction precision over pre-trained approaches. This outcome is valuable for sockpuppet detection, where confidence in positive identifications is paramount. Our approach\footnote{\label{fn}\url{https://github.com/lraszewski/wiki-socks}} is applicable to other online communities as well, and we release it publicly.

\paragraph{2. We construct and publicly release a dataset of sockpuppet investigations on Wikipedia.} Our dataset\footnotemark[\value{footnote}] improves upon existing datasets which are either outdated \citep{solorio2013case}, unreleased publicly \citep{kumar2017} or do not preserve investigation structure \citep{sakib2022}.

\paragraph{3. We formulate a more realistic task definition.} Previous approaches \citep{sakib2022} train on data from any number of accounts within a sockpuppet-group, preemptively revealing any deceptive efforts made by a puppetmaster to the model. Our model is fine-tuned on just one accused user, as it would be when deployed.
\section{Related Work}

\subsection{Sockpuppetry}

\textit{Sockpuppetry} is typically described as the use of multiple accounts by a single user for deceptive or malicious purposes \citep{zheng2011, solorio2013case, bu2013, liu2016, sakib2022}, however, not all sockpuppets are malicious. \citet{kumar2017} provide a more general definition: A \textit{sockpuppet} is any account controlled by a user with at least one other account. The set of these accounts is referred to as a \textit{sockpuppet-group}, and their controlling user their \textit{puppetmaster}. We use this definition with one small amendment: that these accounts be, at some point, operated concurrently. This adjustment distinguishes the task of sockpuppet detection from that of ban evasion, where secondary accounts are created strictly after the primary accounts are banned \citep{niverthi2022}. Other similar tasks include bot campaign detection, where accounts are controlled by automated agents instead of humans. Advancements in human-like text generation is blurring this distinction \citep{sallah2024}.

\subsubsection{Motivation}\label{motivations}

Malicious users use sockpuppets to vandalise Wikipedia pages \citep{solorio2013case}, propagandise political views \citep{kumar2017, afroz2012}, or improve their own public image \citep{owens2013}. Sockpuppets undermine collaboration on Wikipedia through false majority opinions, vote stacking \citep{solorio2013case} and \textit{Straw man socks}, which argue easily refuted opposing arguments to discredit opposition \citep{kumar2017}.

\subsubsection{Detection}

The current approach to detecting sockpuppetry on Wikipedia is manual\footnote{\url{https://en.wikipedia.org/wiki/Wikipedia:Sockpuppet_investigations}}: users argue their case before a presiding administrator, who may supplement evidence with technical logs. Once a verdict is reached, guilty accounts are suspended and the investigation is archived.

Many automated approaches have been proposed to support this process. 
\textit{Authorship Attribution} (AA) has been used to determine whether the intent and writing style of accounts are similar enough to be the same user. Linear classifiers with manually selected authorship features have achieved consistent results \citep{solorio2013case, bu2013, liu2016, sakib2022}. They use lexical, structural and syntactic features \citep{bu2013}. AA classifiers struggle when given only short pieces of text \citep{shrestha2017}, which is typically all that is available in the case of sockpuppet detection \citep{solorio2013case}. Features may also require domain specific selection \citep{kotsiantis2007}, and typically act under the assumption that the authors are not attempting to evade detection \citep{solorio2013case}. Faced with adversarial authors, commonly used authorship features are easily evaded \citep{brennan2012}.

\textit{Meta-data} approaches focus on the behaviour of users. \citet{tsikerdekis2014} identified that the number and time between edits deviates from that of normal users over time. Meta-data approaches typically do not require pairwise comparison between accounts, which reduces computational complexity. \citet{kumar2017} highlight six points of divergence from usual user activity. These features can be combined with authorship features for improved performance \citep{solorio2013case, sakib2022}. In all prior literature, approaches consider a single model that classifies users or edits as belonging to a sockpuppet or not, rather than training a new model for each investigation \citep{solorio2013case, bu2013, liu2016, sakib2022}.

\subsection{Meta-learning}

\textit{Meta-learning} research focuses on the problem of ``learning to learn''. In this setting, a machine learning model gains experience over a collection of tasks, rather than just one, and in doing so improves its performance on future tasks \citep{hospedales2020}. \citet{hospedales2020} define \textit{base-learning} as the inner learning algorithm solving a task, such as authorship attribution. \textit{Meta-learning} is an outer learning algorithm which updates the inner algorithm according to its own \textit{meta-objective}, typically quick adaptation to new tasks \citep{finn2017, snell2017, so2021}.

Meta-learning has been successful in many domains, such as image classification \citep{antoniou2019}, sentiment analysis \citep{liang2023}, and text classification \citep{bansal2021}. \citet{tian2023} investigated the approach to detect state-sponsored trolls. Beyond reducing data dependence, other limitations of deep neural networks, such as unsupervised learning performance, may also be improved \citep{hospedales2020}.

Some limitations include a requirement for a large number of tasks \citep{shedivat2021}, significant compute costs of first-order techniques \citep{finn2017}, generalisation to out of distribution tasks and a lack of realistic benchmark datasets \citep{vettoruzzo2023}.

\paragraph{Gradient-based}\label{gradient-based}

Meta-Learning approaches are made up of several families. Gradient-based approaches use gradient descent to update a model's parameters to minimise the loss according to a meta-objective. These approaches are model-agnostic, making them advantageous over Metric and Model-based approaches that make restrictions on model architecture. The foremost approach is MAML \citep{finn2017} and its successors \citep{antoniou2019, triantafillou2020, finn2018, rajeswaran2019}. A related approach is Reptile \citep{nichol2018} that requires significantly less compute whilst achieving similar performance \citep{vinyals2016}. An advantage of Reptile over MAML is that it does not require a train-test split for each training task \citep{nichol2018}, allowing more data to be used in training.

\paragraph{Metric-based}

Metric based approaches learn a distance function (metric) that clusters samples from the same class together. This metric should then generalise well on new tasks \citep{hospedales2020}, and in most approaches does not require fine-tuning \citep{sung2018}. They include prototypical networks \citep{snell2017}, siamese networks \citep{Koch2015}, relation networks \citep{sung2018} and matching networks \citep{vinyals2016}. Unlike gradient-based approaches, they make some restrictions on model-architecture.

\paragraph{Model-based}

Model-based methods use a model architecture specifically designed for rapid parameter updates. They include extended neural turing machines \citep{santoro2016} and meta-networks \citep{munkhadalai2017}. Their main drawback is their inherent restrictions on model design.
\section{Dataset}
We create and release\footnote{\url{https://github.com/lraszewski/wiki-socks}} a novel sockpuppetry dataset for this study. Using the meta-learning paradigm, each investigation is framed as a discrete problem, where writing samples from a single sockpuppet-group must be separated from writing samples of non-sockpuppets.

Tasks consist of writing samples from both sockpuppet and non-sockpuppet users. Edits to Wikipedia are called \textit{contributions}, and contain a \textit{message} component where the user can describe their edit. As in previous approaches \citep{solorio2013case, sakib2022}, we use these contribution messages as the writing samples. Negative samples are contributions from non-sockpuppet accounts drawn from the same time and article distribution as the sockpuppet-group.

The final dataset consists of $23,610$ tasks. For each contribution, we provide the timestamp, revision ID, ID of the preceding contribution, user name, article title, contribution message, and a binary label. A sample from one investigation is given in Table \ref{tab:sample} in Appendix \ref{app:gen}.

\subsection{Data collection}

To collect the positive samples, the confirmed Wikipedia sockpuppets page\footnote{\url{https://en.wikipedia.org/wiki/Category:Wikipedia_sockpuppets}} was crawled using a combination of Pywikibot\footnote{\url{https://github.com/wikimedia/pywikibot}} and MediaWiki\footnote{\url{https://www.mediawiki.org/wiki/MediaWiki}} API calls, which extracted each investigation page and the contributions of each confirmed sockpuppet.

Negative samples for each task were collected from the same articles and within the active time period (first and last contribution) of the task's sockpuppet-group. For each positive sample, two random timestamps were drawn, and the next ten contributions made after each was collected. From each set of ten, the first valid (non-duplicate, non-sockpuppet) sample was selected. Multiple samples from the set of ten were not collected, so as to maintain a uniformly random temporal distribution. If a set of ten contained no valid negative samples, the attempt was abandoned. This occurred in cases where the active time period was very short, or the articles were inactive or new.

In reality, there is a class imbalance between the number of sockpuppet and non-sockpuppet accounts. Negative contributions were thus over sampled. Arbitrarily, an ideal ratio of two negative samples to each positive was set. It is unclear how an informed estimate might be reached: the true ratio of genuine users to sockpuppets would not only be too extreme to replicate or learn with, but investigations only occur on accused users, not all users. Ideally one might ascertain the ratio of accused sockpuppets to confirmed sockpuppets, however as the investigations of falsely accused accounts are not archived together, this is impractical to obtain.

For investigations that did not reach the ideal ratio of two negatives for each positive, a second identical pass was performed. This strategy over-sampled articles with more non-sockpuppet editors to make up for the shortfall. This yielded some tasks with up to four negatives for each positive.

664 investigations failed to collect any negative samples.  This may have occurred for a few reasons, namely that the sockpuppet activity was contained to pages without any non-sockpuppet editors\footnote{\url{https://en.wikipedia.org/wiki/Category:Wikipedia_sockpuppets_of_Alliasalmon}}, that the sockpuppet activity was confined to a short period of time that did not contain any other editors\footnote{\url{https://en.wikipedia.org/wiki/Category:Wikipedia_sockpuppets_of_Appearedclip}}, or that the sockpuppets made very few contributions\footnote{\url{https://en.wikipedia.org/wiki/Category:Wikipedia_sockpuppets_of_Lolawin}}. These investigations were retained in the dataset, but were excluded from any experiments. 984 investigations contained no positive samples. These were removed, and a list of empty investigations provided alongside the rest of the dataset.
\section{Method}\label{method}

We provide two task definitions. The first is a description of the \textit{base-learning} problem, which considers training and evaluating a classifier on a single task. This is also referred to as the \textit{inner-loop} in the context of meta-learning. The second description is of the meta-learning problem, and describes how a meta-model is learnt across a distribution of base-learning tasks. This is also called the \textit{outer-loop}.

\subsection{Base-learning}

The base-learning task is a binary classification problem. As input, the model will receive two data sources: the article \textit{page} and \textit{message} describing the contribution. The model outputs a classification, identifying the contribution as either a positive (made by a sockpuppet), or negative sample.

In a deployed setting, there is no given list of confirmed sockpuppets, only a set of accused accounts. Therefore, a model may only be trained on the contributions of a single accused user, which may then be tested on the contributions of the remaining accused accounts. We make similar restrictions on our training data: for each investigation, we define the \textit{puppetmaster} as the sockpuppet with the most contributions. A model is then trained on their contributions. This model is assessed by its ability to distinguish the contributions of the remaining sockpuppets from the samples of non-sockpuppets. By contrast, prior works \citep{sakib2022, solorio2013case} train on data from a mix of accounts. If a puppetmaster attempts to change their behaviour on sockpuppet accounts, this may be unrealistically revealed to a model in training.

We call the set of puppetmaster samples the train set, and the set of sockpuppet samples the test set. Negative samples are split between the two sets proportionally. Due to how these sets are created, test sets much larger than the train set are common.

A validation set is split from the train set, containing 20\% of the available samples and maintaining the same proportion of positive and negative samples. This set is used during base-learning to provide feedback as the model is being trained, and to prevent over-fitting via early stopping. We provide summary statistics of the train-test split in Table \ref{tab:train-test-split} in Appendix \ref{app:gen}.

\subsection{Meta-learning}

The meta-learning problem considers a distribution of tasks \citep{finn2017}. This distribution is split into two sets, meta-train and meta-test. The meta-train set is used for the meta-learning process, whilst the meta-test set is used to evaluate how well the meta-learned model performs. The average performance of the models on these tasks is what is reported in Section \ref{results}.

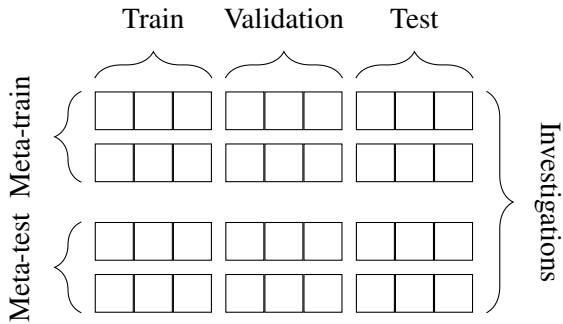
\begin{figure}[t]
    \centering
    \begin{tikzpicture}[node distance=1cm]

        \def\gap{5pt}
        \def\bracedistance{5pt}

        \matrix[
            matrix of nodes,
            nodes={draw, minimum size=0.5cm, anchor=center},
            column sep=0pt, row sep=\gap
        ] (dataset) {
            {} & {} & {} &[\gap] & {} & {} & {} &[\gap] & {} & {} & {} \\
            {} & {} & {} &[\gap] & {} & {} & {} &[\gap] & {} & {} & {} \\
            [\gap] \\
            {} & {} & {} &[\gap] & {} & {} & {} &[\gap] & {} & {} & {} \\
            {} & {} & {} &[\gap] & {} & {} & {} &[\gap] & {} & {} & {} \\
        };
        
        \draw [decorate,decoration={brace,amplitude=10pt}] 
            ([yshift=\bracedistance]dataset-1-1.north west) -- ([yshift=\bracedistance]dataset-1-3.north east) 
            node[midway,above=15pt] (train-set-label) {Train};

        \draw [decorate,decoration={brace,amplitude=10pt}] 
            ([yshift=\bracedistance]dataset-1-5.north west) -- ([yshift=\bracedistance]dataset-1-7.north east) 
            node[midway,above=15pt] (validation-set-label) {Validation};

        \draw [decorate,decoration={brace,amplitude=10pt}] 
            ([yshift=\bracedistance]dataset-1-9.north west) -- ([yshift=\bracedistance]dataset-1-11.north east) 
            node[midway,above=15pt] (test-set-label) {Test};

        \draw [decorate,decoration={brace,amplitude=10pt}] 
            ([xshift=-\bracedistance]dataset-2-1.south west) -- ([xshift=-\bracedistance]dataset-1-1.north west) 
            node[midway,left=5pt] (meta-train-label) {\parbox{1cm}{\centering \rotatebox{90}{Meta-train}}};;

        \draw [decorate,decoration={brace,amplitude=10pt}] 
            ([xshift=-\bracedistance]dataset-5-1.south west) -- ([xshift=-\bracedistance]dataset-4-1.north west) 
            node[midway,left=5pt] (meta-test-label) {\parbox{1cm}{\centering \rotatebox{90}{Meta-test}}};;

        \draw [decorate,decoration={brace,amplitude=10pt}] 
            ([xshift=\bracedistance]dataset-1-11.north east) -- ([xshift=\bracedistance]dataset-5-11.south east) 
            node[midway,right=5pt] (investigations-label) {\parbox{1cm}{\centering \rotatebox{-90}{Investigations}}};;
        
    \end{tikzpicture}
    \caption{Dataset topology.}
    \label{fig:dataset-topology}
\end{figure}

Figure \ref{fig:dataset-topology} depicts the dataset topology, where each row represents the samples of a task. The tasks are split into the meta-train and meta-test sets, and each task is split into train, validation and test sets.

We make several restrictions on the shape of this distribution. To ensure that each task has an over representation of negative samples, we limit the distribution to only include tasks with a negative to positive ratio of at least one. We also ensure that each task has at least ten puppetmaster samples and five sockpuppet samples. These restrictions are derived from the model architecture, which uses a triplet contrastive loss function that requires at least two positive samples in each task. By ensuring each task contains at least ten puppetmaster samples, we guarantee a valid validation set. These restrictions reduce the total number of tasks from $23,610$ to $13,549$. 
90\% of tasks ($12,194$) are reserved for the meta-train set, and the rest ($1,355$) become the meta-test set. To compare the model with non-meta-learning approaches, the meta-train set will either be used for the pre-trained approach, or, where no pre-training is necessary, will not be used at all.

Whilst training on the meta-train set of tasks, approaches are not required to maintain the distinction between task specific train, test and validation sets. These sets are only preserved to the extent that they are required for the meta-learning or pre-trained approach. The training process on each meta-test task is kept constant throughout each approach. Each model is given a maximum of ten epochs to train on the new task before predictions must be made. The metrics of each task in the meta-test set are then averaged to create the overall metrics for the approach. Each approach is run three times, and their mean and standard deviation reported in Section \ref{results}.

\subsection{Meta-learning Strategy}

We use Reptile as our meta-learning strategy. This is because it is similarly performant to MAML whilst being less computationally expensive \citep{nichol2018, rajeswaran2019}. Compared to metric and model-based approaches, Reptile has the benefit of being model-agnostic, allowing flexibility in model architecture.

We use the serial implementation that updates the parameters directly through linear interpolation. It works by repeatedly adapting a clone of the meta-model to a task $\mathcal{T}_i$, and then moving the parameters of the original meta-model $\theta$ toward the adapted parameters $\theta'$ by a factor of $\epsilon$. The parameters of the meta-model therefore move in a direction common to the optimal parameters of each task. The Reptile algorithm is provided in Algorithm \ref{alg:reptile}.

\begin{algorithm}
\caption{Reptile Algorithm (Serial)}
\label{alg:reptile}
\begin{algorithmic}
    \STATE \textbf{Input:} Interpolation rate $\epsilon$, inner steps $k$, task distribution $p(\mathcal{T})$, initial model parameters $\theta$
    \FOR{iteration $=1,2,\dots$}
        \STATE Sample a task $\mathcal{T}_i \in p(\mathcal{T})$
        \STATE Compute $\theta'=U_{\mathcal{T}_i}^k(\theta)$, denoting $k$ steps of SGD or Adam
        \STATE Update $\theta\leftarrow\theta+\epsilon(\theta'-\theta)$
    \ENDFOR
\end{algorithmic}
\end{algorithm}

\section{Model Architecture}

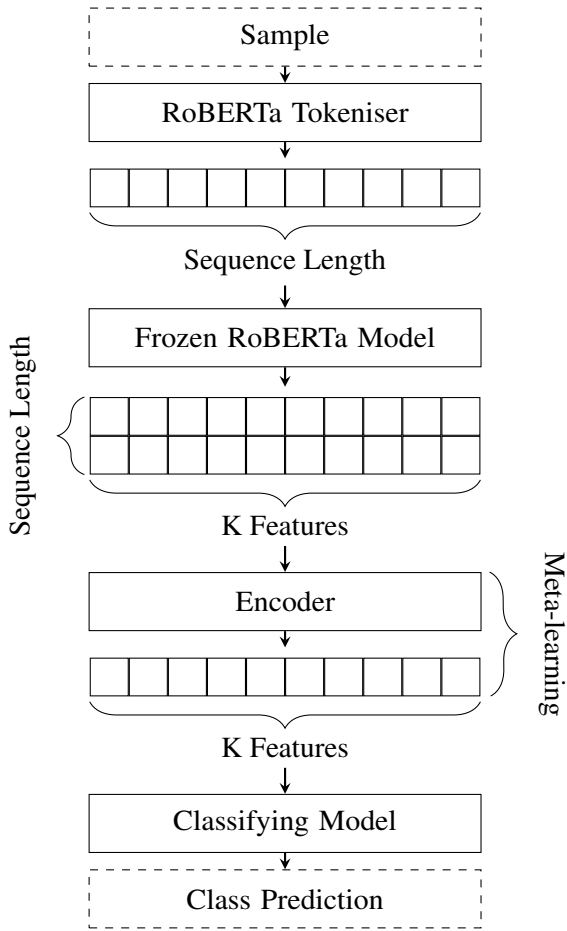
\begin{figure}[ht]
    \centering
    \begin{tikzpicture}[node distance=1cm]

        \node (sample) [block, dashed] {Sample};
        \node (tokenizer) [block, below of=sample] {RoBERTa Tokeniser};

        \matrix[matrix of nodes, below of=tokenizer,
            nodes={draw, minimum size=0.5cm, anchor=center},
            column sep=0pt, row sep=0pt
        ] (tokenized) {
            {} & {} & {} & {} & {} & {} & {} & {} & {} & {} \\
        };
        \draw [decorate,decoration={brace,amplitude=10pt,mirror}] 
            ([yshift=-2pt]tokenized-1-1.south west) -- ([yshift=-2pt]tokenized-1-10.south east) 
            node[midway,below=10pt] (sequence-length) {Sequence Length};
        
        \node (roberta) [block, below of=sequence-length] {Frozen RoBERTa Model};

        \matrix[matrix of nodes, below of=roberta, node distance=1.3cm,
            nodes={draw, minimum size=0.5cm, anchor=center},
            column sep=0pt, row sep=0pt
        ] (roberta-embedding) {
            {} & {} & {} & {} & {} & {} & {} & {} & {} & {} \\
            {} & {} & {} & {} & {} & {} & {} & {} & {} & {} \\
        };
        \draw [decorate,decoration={brace,amplitude=10pt,mirror}] 
            ([yshift=-2pt]roberta-embedding-2-1.south west) -- ([yshift=-2pt]roberta-embedding-2-10.south east) 
            node[midway,below=10pt] (features-label-1) {K Features}; 
        \draw [decorate,decoration={brace,amplitude=10pt}] 
            ([xshift=-2pt]roberta-embedding-2-1.south west) -- ([xshift=-2pt]roberta-embedding-1-1.north west) 
            node[midway, left=5pt] {\parbox{1cm}{\centering \rotatebox{90}{Sequence Length}}};

        \node (model) [below of=features-label-1, block] {Encoder};
        
        \matrix[matrix of nodes, below of=model,
            nodes={draw, minimum size=0.5cm, anchor=center},
            column sep=0pt, row sep=0pt
        ] (model-embedding) {
            {} & {} & {} & {} & {} & {} & {} & {} & {} & {} \\
        };
        \draw [decorate,decoration={brace,amplitude=10pt,mirror}] 
            ([yshift=-2pt]model-embedding-1-1.south west) -- ([yshift=-2pt]model-embedding-1-10.south east) 
            node[midway,below=10pt] (features-label-2) {K Features};

        \draw [decorate,decoration={brace,amplitude=10pt}] 
        ([xshift=3pt]model.north east) -- ([xshift=0pt, yshift=4pt]model-embedding.south east)
        node[midway, right=5pt] {\parbox{1cm}{\centering \rotatebox{-90}{Meta-learning}}};

        \node (classifier) [below of=features-label-2, block] {Classifying Model};
        \node (prediction) [below of=classifier, block, dashed] {Class Prediction};

        \draw [arrow] (sample) -- (tokenizer);
        \draw [arrow] (tokenizer) -- (tokenized);
        \draw [arrow] (sequence-length) -- (roberta);
        \draw [arrow] (roberta) -- (roberta-embedding);
        \draw [arrow] (features-label-1) -- (model);
        \draw [arrow] (model) -- (model-embedding);
        \draw [arrow] (features-label-2) -- (classifier);
        \draw [arrow] (classifier) -- (prediction);
        
    \end{tikzpicture}
    \caption{Model topology.}
    \label{fig:model-diagram}
\end{figure}

A diagram of our approach is given in Figure \ref{fig:model-diagram}. The sample represents a positive or negative contribution. The two textual inputs of the contribution, \textit{page} and \textit{message}, are concatenated using the appropriate separator token. We use RoBERTa\footnote{\url{https://huggingface.co/sentence-transformers/all-distilroberta-v1}} \citep{liu2019}, a pre-trained transformer with frozen parameters, to generate a matrix of contextualised word embeddings. This approach is typical of recent authorship attribution classifiers \citep{delanghe2024, huertastato2022, ai2022, rivera-soto2021}.

This matrix is then fed to a transformer encoder, optimised using Adam \citep{diederik2014}. This encoder is where the majority of task learning takes place, and is the model that will be trained using meta-learning.

Once the authorship embeddings are produced, a neural network processes the embedding to produce a logit, which is later transformed to produce discrete classifications. The classifier has two fully connected layers of dimension $768$ with dropout. It also uses the Adam optimiser, and is trained using a cross-entropy loss function. The classifier is trained on the embeddings produced after the encoder has been trained.

\subsection{Loss Functions}\label{loss-functions}

The encoder model is trained using triplet margin loss \citep{schroff2015}. A contrastive loss function is typical in natural language tasks comparing document similarity \citep{pennington2014, devlin2019}, and also has prior success in contrastive authorship models \citep{huertastato2022}. The triplets are created by iterating through all the positive samples as the anchor, and randomly selecting another negative and positive. Typically, the anchor may be drawn from both classes, however we limit triplets to the positive sockpuppet class. This is because the authors in the negative samples are all different, and therefore should occupy different regions in the embedding space -- only positive samples should be clustered together. 
For the classifier models that interpret the embeddings, we use binary cross-entropy loss.

\subsection{Training Parameters}\label{training-parameters}

When training both the classifier and encoder on the meta-test set of tasks, each was given a maximum of ten epochs, where each epoch is one complete pass through the train set. This is a limitation of the time and computing resources available.

We used a variable batch size strategy that scaled with the length of the task. This catered for smaller tasks whilst still allowing larger tasks to benefit from the stability and speed of larger batch sizes. We used early stopping based on the validation loss with a patience of 3 epochs. During the meta-learning stage, the model was trained over five epochs of the meta-train set of tasks. On each task, Reptile performed five gradient steps before the parameters were updated using an interpolation rate of $0.2$. The $\beta_1$ parameter of Adam was set to 0 as recommended \citep{nichol2018}. At the end of each epoch, the model was saved along with the sum of the training loss of each task in that epoch. The best performing model relative to the validation loss was selected.

\subsection{Hyper-parameters}\label{hyper-parameters}

We tuned model hyper-parameters using the Optuna\footnote{\url{https://optuna.org/}}
framework. The encoder and classifier models were tuned together. 100 tests were run, where the model was trained over ten randomly selected tasks from the meta-train distribution of tasks. The performance of each on their respective tasks was averaged and provided to the optimiser as feedback. We performed three optimisations, one for the encoder and classifier models, a second for the RoBERTa baseline classifier (Section \ref{baselines}), and a third for the Reptile parameters. We provide the tuned hyper-parameters in Table \ref{tab:hyper-parameters} in Appendix \ref{app:gen}.

\section{Metrics}\label{metrics}

The main metrics for comparison between approaches should be the area-based metrics, AUROC and AUPRC. This is because they do not require a specific threshold to be decided, which may distort the appearance of classifier performance. We also provide the F1-Score and F0.5-Score. The first is justified by previous literature \citep{sakib2022, solorio2013case, solorio2013sockpuppet}, whilst the second presents a balance between recall and precision more relevant to the deployment environment, where false positives are strongly discouraged\footnote{\url{https://en.wikipedia.org/w/index.php?title=Wikipedia:Sockpuppet_investigations/SPI/Administrators_instructions&oldid=1173289303}}. We also provide the accuracy, precision, and recall of each model as supplementary metrics.

Reported metrics undergo an aggregation process. For each experiment, the results of the approach on the test set of each task in the meta-test set (see Figure \ref{fig:dataset-topology}) are computed. The results of each task are then averaged to find the overall result of the approach for that experiment. Three experiments of each approach are run. The metrics across each experiment are averaged, and the standard deviation provided as a confidence interval.

\subsection{Baselines}\label{baselines}

We consider several baselines, including two trivial baselines (random and majority classifiers), also used by \citet{solorio2013case}. In the case of the majority baseline, the class predicted is based on the training dataset.

\paragraph{RoBERTa baseline}

To assess whether the encoder model itself provides a significant improvement, we train a simple binary classifier on the sentence-level RoBERTa embeddings. This changes the model architecture by reducing the output from the frozen RoBERTa model from a two-dimensional matrix to a one-dimensional vector. The vector is then fed directly into a fully connected neural network classifier. \citet{huertastato2022} employed a similar baseline to assess their authorship representation learner.

\paragraph{Non-meta-learning approach}

To isolate the effects of meta-learning, we also test our approach without it. This model will follow the same training approach as the Meta-learned model on test tasks.

\paragraph{Pre-trained approach}

The pre-trained approach trains our model on a merged dataset of the meta-train set of tasks. This is the approach used in prior literature \citep{solorio2013case, sakib2022}. At test time, this approach will be fine-tuned on tasks in the meta-test set.

\paragraph{Upper limit}

As a significant portion of all contributions do not have any text in their \textit{message} feature (24.45\% of all contributions, 63.64\% of which are positive samples), these contributions are indistinguishable from one another using the provided features. Therefore, the upper limit of performance is more accurately defined as the perfect classifier on all contributions where a \textit{message} value is present, and a random classifier otherwise.
\section{Results}\label{results}

\begin{table*}
  \centering
  \resizebox{\textwidth}{!}{%
  \begin{tabular}{lccccccc}
    \hline
    \textbf{Approach} & \textbf{AUROC} & \textbf{AUPRC} & \textbf{F1-Score} & \textbf{F0.5-Score} & \textbf{Accuracy} & \textbf{Precision} & \textbf{Recall} \\
    \hline
    Random & $50.10\pm0.14$ & $50.85\pm0.09$ & $40.34\pm0.11$ & $36.52\pm0.12$ & $50.10\pm0.16$ & $34.46\pm0.12$ & $50.05\pm0.15$ \\
    Majority & - & - & - & - & $65.60\pm0.00$ & - & - \\
    RoBERTa & $65.70\pm0.00$ & $50.45\pm0.03$ & $57.97\pm0.01$ & $57.52\pm0.06$ & $66.98\pm0.06$ & $59.54\pm0.13$ & $67.63\pm0.17$ \\
    Standard Enc. & $68.33\pm0.09$ & $50.67\pm0.33$ & $60.05\pm0.18$ & $58.73\pm0.16$ & $68.90\pm0.07$ & $59.72\pm0.32$ & $69.88\pm0.34$ \\
    Pre-trained Enc. & $62.74\pm0.02$ & $44.80\pm0.19$ & $57.49\pm0.13$ & $52.90\pm0.12$ & $62.79\pm0.05$ & $51.45\pm0.25$ & $74.76\pm0.28$ \\
    Reptile Enc. & $\mathbf{78.98\pm0.12}$* & $\mathbf{62.21\pm0.08}$* & $\mathbf{67.46\pm0.53}$* & $\mathbf{67.89\pm0.17}$* & $\mathbf{77.51\pm0.19}$* & $\mathbf{69.43\pm0.26}$* & $\mathbf{70.81\pm0.82}$ \\
    \hline
    Upper Limit & $96.73\pm0.00$ & $93.56\pm0.00$ & $86.48\pm0.00$ & $91.11\pm0.00$ & $92.01\pm0.00$ & $95.38\pm0.00$ & $81.66\pm0.00$ \\
    \hline
  \end{tabular}
  }
  \caption{Results for the sockpuppet prediction task. Asterisks indicate statistical significance compared to the standard encoder (p < 0.05).}
  \label{tab:results}
\end{table*}

Our results are presented in Table \ref{tab:results}. Metrics are measured as the mean across all three test runs with the standard deviation as error bounds. The classification threshold used to compute predictions from logits for applicable metrics were computed at a task level, using the optimal threshold relative to the F0.5-Score on the task validation set. A T-test was used to evaluate the statistical significance of the meta-encoder compared to the standard encoder, indicated with asterisks. The averaged ROC and PR curves are provided in Appendix \ref{app:gen}.

The meta-learning approach significantly outperforms other approaches ($P<<0.05$) in AUROC, AUPRC, F1-score, F0.5-score and accuracy, with substantial improvements of approximately $10\%$. Recall did not improve significantly, tying the overall improvement to an increase in precision. This suggests that meta-learning helps the classifier make fewer false positive predictions whilst preserving its ability to identify true positives. This result is desirable for the detection of any malicious behaviour, as unjustly punishing innocent users is typically considered worse than missing a few malicious ones -- the emphasis is on a very high confidence in positive predictions. 

The metrics of the pre-trained encoder are unexpected. The additional pre-training should have provided the model with a general understanding of the task prior to fine-tuning, however the approach falls behind the non-pre-trained encoder model on most metrics. In prior work \citep{sakib2022, solorio2013case}, the pre-trained approach performed well. This reduction in performance may be due to the harder task setting, however there may be other causes. \citet{sakib2022} combined their authorship attribution features with behavioural features, which may be more consistent across tasks, and therefore better for the pre-trained approach. This suggests approaches that focus on authorship attribution may require a model to have greater adaptability.

The lowest performing metric is the AUPRC result. This is likely due to the class imbalance, which overall consisted of $65.60\%$ negative and $34.40\%$ positive samples. The AUPRC is more sensitive to `hard' negative samples, as precision decreases substantially for each false positive.

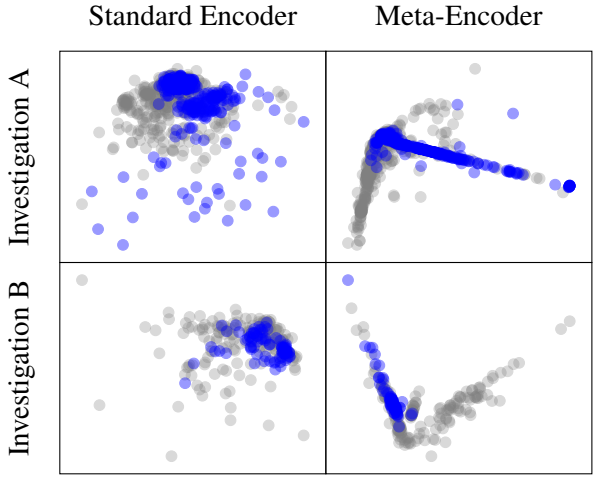
\begin{figure}[t]
    \centering
    \begin{tikzpicture}
    \begin{groupplot}[
        group style={group size=2 by 2, horizontal sep=0em, vertical sep=0em},
        scatter,
        point meta=explicit,
        every axis plot/.style={only marks, opacity=0.4},
        scatter/classes={
            1={blue},
            0={gray, opacity=0.3}
        },
        xtick=\empty,
        ytick=\empty,
        ylabel style={yshift=0.5em},
        width=0.66\linewidth,
    ]
        \nextgroupplot[title={Standard Encoder}, ylabel={Investigation A}]
        \addplot table[x=pc1, y=pc2, col sep=comma, meta=label] {data/Category_Wikipedia_sockpuppets_of_Film_Fan_pca_dual_baseline_1.csv};
        \nextgroupplot[title={Meta-Encoder}]
        \addplot table[x=pc1, y=pc2, col sep=comma, meta=label] {data/Category_Wikipedia_sockpuppets_of_Film_Fan_pca_dual_reptile_1.csv};
        \nextgroupplot[ylabel={Investigation B}]
        \addplot table[x=pc1, y=pc2, col sep=comma, meta=label] {data/Category_Wikipedia_sockpuppets_of_Al_aman_kollam_pca_dual_baseline_1.csv};
        \nextgroupplot[]
        \addplot table[x=pc1, y=pc2, col sep=comma, meta=label] {data/Category_Wikipedia_sockpuppets_of_Al_aman_kollam_pca_dual_reptile_1.csv};
    \end{groupplot}
    \end{tikzpicture}
    \caption{PCA of high performing embeddings. Blue circles represent positive samples.}
    \label{fig:pcas-good}
\end{figure}

\begin{figure}[t]
    \centering
    \begin{tikzpicture}
    \begin{groupplot}[
        group style={group size=2 by 2, horizontal sep=0em, vertical sep=0em},
        scatter,
        point meta=explicit,
        every axis plot/.style={only marks, opacity=0.4, mark options={draw=black}},
        scatter/classes={
            1={blue},
            0={gray, opacity=0.3}
        },
        xtick=\empty,
        ytick=\empty,
        ylabel style={yshift=0.5em},
        width=0.66\linewidth,
    ]
        \nextgroupplot[title={Standard Encoder}, ylabel={Investigation C}]
        \addplot table[x=pc1, y=pc2, col sep=comma, meta=label] {data/Category_Wikipedia_sockpuppets_of_Amirharbo_pca_dual_baseline_1.csv};
        \nextgroupplot[title={Meta-Encoder}]
        \addplot table[x=pc1, y=pc2, col sep=comma, meta=label] {data/Category_Wikipedia_sockpuppets_of_Amirharbo_pca_dual_reptile_1.csv};
        \nextgroupplot[ylabel={Investigation D}]
        \addplot table[x=pc1, y=pc2, col sep=comma, meta=label] {data/Category_Wikipedia_sockpuppets_of_Cameronfree_pca_dual_baseline_1.csv};
        \nextgroupplot[]
        \addplot table[x=pc1, y=pc2, col sep=comma, meta=label] {data/Category_Wikipedia_sockpuppets_of_Cameronfree_pca_dual_reptile_1.csv};
    \end{groupplot}
    \end{tikzpicture}
    \caption{PCA of low performing embeddings. Blue circles represent positive samples.}
    \label{fig:pcas-bad}
\end{figure}
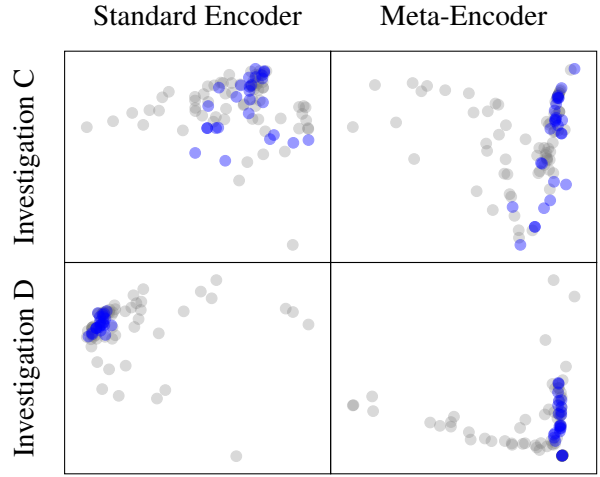

This result suggests that despite an increase in the precision of predictions being the principal benefit of meta-learning, it still remains the model's main flaw. This conclusion is further corroborated in Section \ref{error-analysis}. Intriguingly, both non-meta-learning approaches achieved marginally worse scores than the random baseline. The similarly low precision scores corroborate the earlier statement that the principal improvement of meta-learning in this domain is the reduction in false positives.

Surprisingly, the performance difference between the basic RoBERTa classifier and the encoder model is small. The encoder model was expected to perform better as it is trained on the word level embeddings produced by RoBERTa, and therefore should have had a richer understanding of user writing style than the semantic sentence level embeddings used in the RoBERTa classifier. In all metrics the encoder performs slightly better, suggesting there is some truth to the hypothesis, however, the small training set sizes may have prevented a significant divergence.

Whilst prior Wikipedia sockpuppet detection approaches report higher F1-scores of 73 \citep{solorio2013case} and 82 \citep{sakib2022}, the difference in datasets and task construction (neither study distinguishes between sockpuppets and puppetmasters) make a fair comparison difficult.

\subsection{Error Analysis}\label{error-analysis}

Figure \ref{fig:pcas-good} provides insight into the effect of meta-learning on the embeddings. The embeddings of two test investigations\footnote{Investigations of \textit{Film\_Fan} (A) and \textit{Al\_aman\_kollam} (B).} have been projected to two dimensions using Principal Component Analysis (PCA), with positive samples being coloured in blue. The left-hand column contains the embeddings of the test samples produced by the standard encoder model after training on the task. The right-hand column are the embeddings produced with the meta-encoder.

\begin{table*}[t]
    \centering
    \begin{tabularx}{\textwidth}{XXXXX}
        \hline
        \textbf{Positive Samples} & \textbf{RoBERTa} & \textbf{Standard Enc.} & \textbf{Pre-trained Enc.} & \textbf{Reptile Enc.} \\
        \hline
        $[10,20)$ & $64.66\pm0.26$ & $66.65\pm0.55$ & $64.23\pm0.45$ & $\mathbf{79.37\pm0.99}$* \\
        $[20,30)$ & $64.48\pm0.22$ & $67.32\pm0.45$ & $64.74\pm0.33$ & $\mathbf{80.00\pm0.02}$* \\
        $[30,40)$ & $63.83\pm0.31$ & $66.96\pm0.38$ & $64.56\pm0.28$ & $\mathbf{79.01\pm0.37}$* \\
        $[40,50)$ & $67.81\pm0.52$ & $70.51\pm0.41$ & $60.79\pm0.34$ & $\mathbf{79.81\pm0.22}$* \\
        $[50, \infty)$ & $66.05\pm0.04$ & $68.73\pm0.15$ & $62.20\pm0.12$ & $\mathbf{78.60\pm0.07}$* \\
        \hline
    \end{tabularx}
    \caption{AUROC scores of approaches on small tasks. Tasks are binned by the number of positive samples. Asterisks indicate statistical significance compared to the standard encoder (p < 0.05).}
    \label{tab:binned-results}
\end{table*}

The embeddings learnt by the meta-encoder appear tighter, with less overlap between the clusters. This is supported by the results of these particular investigations: investigation A received an AUROC of 62\% with the standard encoder, which improved to 83\% using the meta-encoder. Investigation B had a similar improvement, from 69\% to 91\%. Both standard and meta-encoders were able to cluster positive samples together, however the meta-encoder exhibits better separation from negative samples. This aligns with the overall results, where the meta-encoder saw small improvements in recall, but large improvements in precision.

To contrast the successful examples, Figure \ref{fig:pcas-bad} presents two investigations\footnote{Investigations of \textit{Amirharbo} (C) and \textit{Cameronfree} (D).} that performed poorly. Investigation C achieved AUROCs of 52\% (standard encoder) and 54\% (meta-encoder). Investigation D was similar, with a small improvement from 58\% to 61\%. Whilst less defined, the right-angle structure is still evident, and positive samples are still clustered within a single arm, suggesting the encoder has no issues identifying positive samples. The difference then is the proportion of negative samples that appear in the `positive' arm. This again aligns with the overall results, where recall is largely consistent between approaches, and most of the improvement is in the precision of positive classifications. In these two cases, the poor AUROC performance can be attributed to the failure of the meta-encoder to improve upon the precision.

In Investigation D, contribution messages are characteristically short, typically the name of the article section edited. The messages of many negative samples are similar. This convention is easily detected, explaining why both classifiers were able to cluster the positive samples, but found distinguishing them from negative samples using the convention difficult. This may explain why the recall is acceptable, but precision is low. The sockpuppet-group in investigation C use a message of just a couple of words at most, but typically use no message at all. As most Wikipedians add a contribution message (only 8.89\% of negative samples collected had empty message fields, compared to 15.56\% of positive samples), a consistently empty one would identify the sockpuppet to the encoder, but would be indistinguishable from legitimate message-less contributions. This leads to the following conclusion: where a sockpuppet's behaviour is characterised by empty or conventional messages, positive recall is strong, but precision suffers.

\subsection{Small tasks}

Table \ref{tab:binned-results} contains additional binned results for each approach on smaller tasks, containing between ten and fifty positive samples each. These results indicate that the performance increase of our model over other approaches tested holds in tasks with exceptionally low samples available.
\section{Conclusion}

We study the problem of detecting malicious sockpuppetry on Wikipedia. We are the first to propose meta-learning to address the data-scarcity challenge in detecting sockpuppet accounts through writing style. Our results demonstrate significant performance improvements when compared to pre-trained approaches, especially in prediction precision. We attribute this to our approach's ability to quickly adapt to distinct authorship styles with limited samples. In doing so, we defined a more realistic task definition that provides a more accurate measure of performance, and released an updated, verifiable and adaptable dataset of sockpuppet investigations appropriate for future meta-learning research. Our findings extend to any online social platform where users engage in sockpuppetry.

\section*{Limitations}

There are several limitations of our model that could benefit from further research.

As discussed in Section \ref{error-analysis}, our model is limited in cases where sockpuppet contributions contain little or no message data. In these cases, the encoder requires additional information. One way to address this would be to include the edit data of contributions, that is the changes made to the article itself. This would allow a model to understand the intent and implications of a contribution even when a description is absent. A contribution must make changes to the article, and edits themselves are likely to be far more diverse in nature than the messages, providing a model with a strong distinguishing signal. The additional signal would also further improve high performing investigations.

Another limitation is in safety. There are several legitimate reasons why a user might have several accounts. One of these reasons may be for the safety of editors editing politically contentious articles\footnote{\url{https://en.wikipedia.org/wiki/List_of_people_imprisoned_for_editing_Wikipedia}}. Whilst our approach was trained solely on malicious examples of sockpuppetry, no efforts were made to ensure this approach could not reveal benign sockpuppet-groups by mistake. Additional work may focus on providing a safeguard measure that ensures the sockpuppet behaviour being observed is malicious.

Our approach should also be evaluated against Generative language models, which are becoming increasingly effective at creating text that looks human \citep{liu2023}. It is likely that future sockpuppet-groups might utilise generative models to edit Wikipedia, rewriting edits to conceal writing style or to fully automate contributions. Many approaches are already focusing on the detection of text generated by prolific models \citep{dhaini2023}. Future work may evaluate how robust the meta-learning approach is to authorship obfuscation using generative models.

Considering the performance of the approach, it is unable to replace human-led sockpuppet investigations. When found to be guilty of sockpuppetry, accounts are blocked. Some users assign great value to their accounts, and incorrect sockpuppet classifications would be damaging to the community. The approach could serve as an additional source of evidence in open investigations, or as a detection method that triggers human-led investigation on suspicious accounts. To occupy a larger role in investigations, the precision of the approach must improve further.
\section*{Ethical Considerations}

There is a valid concern for privacy when releasing this dataset. Usernames are important to Wikipedia editors \citep{garmager2025}, and may be used to represent a person's real identity, contain some personally identifiable information, or obscure their identity completely.

Arguments against anonymisation are numerous. Whilst this study does not use username data, previous approaches have \citep{sakib2022}, and future approaches may too. As all the data collected is publicly available, any anonymisation attempts would be circumventable. We commit to removing any user who requests removal from our dataset, but maintain the username is a valuable data point for future work. For these reasons, the data was not anonymised.
\bibliography{bibliography}

\appendix
\section{Appendix}\label{app:gen}

\begin{table}[h]
    \centering
    \begin{tabularx}{\columnwidth}{Xc}
        \hline
        \textbf{Statistic} & \textbf{Value} \\
        \hline
        Meta-Train Size & $12194$ \\
        Meta-Test Size & $1355$ \\
        Ave. Train Size & $534.35$ \\
        Ave. Validation Size & $133.58$ \\
        Ave. Test Size & $923.75$ \\
        Ave. Train Positives & $174.71$ \\
        Ave. Train Negatives & $359.63$ \\
        Ave. Validation Positives & $43.67$ \\
        Ave. Validation Negatives & $89.91$ \\
        Ave. Test Positives & $298.74$ \\
        Ave. Test Negatives & $625.01$ \\
        \hline
    \end{tabularx}
    \caption{Train-Test Split statistics.}
    \label{tab:train-test-split}
\end{table}

\begin{table}[h]
    \centering
    \begin{tabularx}{\columnwidth}{Xc}
        \hline
        \textbf{Statistic} & \textbf{Value} \\
        \hline
        Num. Investigations & $23160$ \\
        Ave. Length (Contributions) & $970.39$ \\
        Ave. Positive Samples & $321.47$ \\
        Ave. Negative Samples & $648.91$ \\
        Ave. Puppetmaster Samples & $144.27$ \\
        Ave. Sockpuppet Samples & $177.20$ \\
        Ave. Message Length (Char) & $44.92$ \\
        \hline
    \end{tabularx}
    \caption{Summary Statistics of our dataset.}
    \label{tab:dataset-statistics}
\end{table}

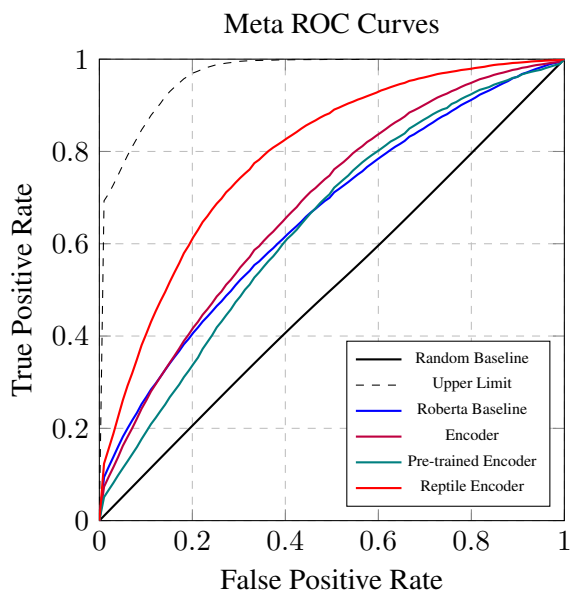
\begin{figure}
    \centering
    \begin{tikzpicture}
        \begin{axis}[
            width=\linewidth, height=\linewidth,
            ymajorgrids=true,
            xmajorgrids=true,
            grid=both,
            grid style=dashed,
            title={Meta ROC Curves},
            xlabel={False Positive Rate},
            ylabel={True Positive Rate},
            xmin=0, xmax=1,
            ymin=0, ymax=1,
            legend pos=south east,
            legend style={font=\tiny},
            axis equal image,
            axis line style={-}
        ]

        \addplot[thick, smooth, black] table [x=fpr,y=tpr,col sep=comma] {data/random_baseline_roc_curve.csv};
        \addlegendentry{Random Baseline}
        

        \addplot[dashed] table [x=fpr,y=tpr,col sep=comma] {data/optimal_baseline_roc_curve.csv};
        \addlegendentry{Upper Limit}
    
        \addplot[thick, blue] table [x=fpr,y=tpr,col sep=comma] {data/roberta_baseline_roc_curve.csv};
        \addlegendentry{Roberta Baseline}

        \addplot[thick, purple] table [x=fpr,y=tpr,col sep=comma] {data/dual_baseline_roc_curve.csv};
        \addlegendentry{Encoder}

        \addplot[thick, teal] table [x=fpr,y=tpr,col sep=comma] {data/dual_pretrained_roc_curve.csv};
        \addlegendentry{Pre-trained Encoder}

        \addplot[thick, red] table [x=fpr,y=tpr,col sep=comma] {data/dual_reptile_roc_curve.csv};
        \addlegendentry{Reptile Encoder}
    
        \end{axis}
    \end{tikzpicture}
    \caption{Average ROC curves of models on test tasks.}
    \label{fig:roc-curves}
\end{figure}

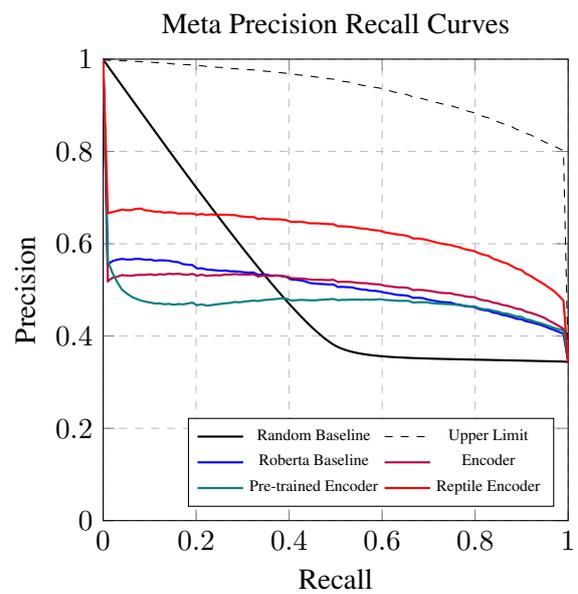
\begin{figure}
    \centering
    \begin{tikzpicture}
        \begin{axis}[
            width=\linewidth, height=\linewidth,
            ymajorgrids=true,
            xmajorgrids=true,
            grid=both,
            grid style=dashed,
            title={Meta Precision Recall Curves},
            xlabel={Recall},
            ylabel={Precision},
            xmin=0, xmax=1,
            ymin=0, ymax=1,
            legend pos=south east,
            legend columns=2,
            legend style={font=\tiny},
            axis equal image,
            axis line style={-}
        ]

        \addplot[thick, smooth, black] table [x=recall,y=precision,col sep=comma] {data/random_baseline_prc_curve.csv};
        \addlegendentry{Random Baseline}


        \addplot[dashed] table [x=recall,y=precision,col sep=comma] {data/optimal_baseline_prc_curve.csv};
        \addlegendentry{Upper Limit}

        \addplot[thick, blue] table [x=recall,y=precision,col sep=comma] {data/roberta_baseline_prc_curve.csv};
        \addlegendentry{Roberta Baseline}

        \addplot[thick, purple] table [x=recall,y=precision,col sep=comma] {data/dual_baseline_prc_curve.csv};
        \addlegendentry{Encoder}

        \addplot[thick, teal] table [x=recall,y=precision,col sep=comma] {data/dual_pretrained_prc_curve.csv};
        \addlegendentry{Pre-trained Encoder}

        \addplot[thick, red] table [x=recall,y=precision,col sep=comma] {data/dual_reptile_prc_curve.csv};
        \addlegendentry{Reptile Encoder}
    
        \end{axis}
    \end{tikzpicture}
    \caption{Average PR curves of models on test tasks.}
    \label{fig:pr-curves}
\end{figure}

\begin{table*}[t]
    \centering
    \begin{tabularx}{\textwidth}{XXc}
        \hline
        \textbf{Model} & \textbf{Hyper-parameter} & \textbf{Value} \\
        \hline
        \multirow{5}{*}{Encoder Model} & Number of Attention Heads & 2              \\
                                       & Number of Layers          & 6              \\
                                       & Learning Rate             & 0.0001         \\
                                       & Loss Margin               & 0.2            \\
                                       & Optimiser                 & Adam           \\
        \hline
        \multirow{5}{*}{Classification Model} & Dropout Chance            & 0.35           \\
                                       & Learning Rate             & 0.001          \\
                                       & Layer 0 Nodes             & 768            \\
                                       & Layer 1 Nodes             & 768            \\
                                       & Optimiser                 & Adam           \\
        \hline
        \multirow{9}{*}{RoBERTa Classifier} & Dropout Chance            & 0.7615         \\
                     & Learning Rate             & 0.0008         \\
                     & Layer 0 Nodes             & 768            \\
                     & Layer 1 Nodes             & 512            \\
                     & Layer 2 Nodes             & 512            \\
                     & Layer 3 Nodes             & 256            \\
                     & Layer 4 Nodes             & 256            \\
                     & Layer 5 Nodes             & 128            \\
                     & Optimiser                 & Adam           \\
        \hline
        \multirow{2}{*}{Reptile} & Interpolation Rate        & 0.2            \\
                     & Number of Steps           & 5 \\
        \hline
    \end{tabularx}
    \caption{Tuned model hyper-parameters.}
    \label{tab:hyper-parameters}
\end{table*}

\begin{table*}
    \centering
    \resizebox{\textwidth}{!}{%
    \begin{tabular}{lllclll}
        \hline
        \textbf{timestamp} & \textbf{revid} & \textbf{parentid} & \textbf{sock} & \textbf{user} & \textbf{page} & \textbf{message} \\
        \hline
        2022-03-15T23:45:35+00:00 & 1077368891 & 1077368777 & 1 & user1 & IShowSpeed & fixed errors \\
        2022-03-15T23:44:47+00:00 & 1077368777 & 1077368713 & 1 & user2 & IShowSpeed & Added Michigan \\
        2022-04-28T20:45:59+00:00 & 1085166027 & 1085165878 & 1 & user3 & Cyclone Batsirai & Where's the source?? \\
        2020-10-01T00:44:54+00:00 & 981220197 & 981144214 & 0 & user4 & User talk:Ohnoitsjamie & /* William Stickman IV partisan editing */ \\
        2019-04-22T01:46:29+00:00 & 893534911 & 892677927 & 0 & user5 & User talk:TheresNoTime & /* Come back! */ new section \\ \hline
    \end{tabular}%
    }
    \caption{Data sample from one investigation.}
    \label{tab:sample}
\end{table*}

\section{NLP For Wikipedia Workshop}

\subsection*{How does this work support the Wikimedia community?}

This work supports the Wikimedia community by investigating a novel approach to sockpuppet detection that can help reduce the burden of manual administration. Our publicly released dataset makes future research quicker, easier and directly comparable.

\subsection*{What license are you using for your data, code, models? Are they available for community re-use?}

We will release our dataset and code under the GPL-3.0 Licence. Both will be available for community reuse, and we encourage further extensions to our dataset.

\subsection*{Did you provide clear descriptions and rationale for any filtering that you applied to your data? For example, did you filter to just one language (e.g., English Wikipedia) or many? Did you filter to any specific geographies or topics?}

We limited our research to English Wikipedia as it has the largest supporting body of prior work and a large accessible record of confirmed sockpuppets.

\subsection*{If there are risks from your work, do any of them apply specifically to Wikimedia editors or the projects?}

As discussed in the Limitations section, despite training solely on malicious sockpuppet-groups, our approach has the potential to reveal benign sockpuppet-groups, either intentionally or by mistake. Wikipedia permits multiple accounts for users whose edits to articles may expose them to political, religious or other forms of persecution. Revealing connections between public and private accounts could put such editors at risk.

\subsection*{Did you name any Wikimedia editors (including username) or provide information exposing an editor's identity?}

Our dataset contains the usernames of confirmed sockpuppets as well as standard users. No additional information about these users is provided. We preserve username data because it has been used in prior sockpuppet detection methods, and we suspect future approaches may do so as well.

\subsection*{Could your research be used to infer sensitive data about individual editors? If so, please explain further.}

All the collected data is publicly available. Provided a very accurate authorship classifier was trained, a bad actor may be able to identify authors outside of Wikipedia, however such a search would be infeasible without an already significantly reduced search space. 

\end{document}